\documentclass{article}
\pdfoutput=1
\usepackage[final]{nips_2017}
\usepackage{times}
\usepackage{graphicx}
\usepackage{subfigure} 
\usepackage{amsmath}
\usepackage{amssymb}
\usepackage{natbib}
\usepackage{algorithm}
\usepackage{algorithmic}
\usepackage{hyperref}

\usepackage{comment}
\usepackage{bm}
\usepackage{amsmath}
\usepackage{amssymb}
\usepackage{amsfonts}
\usepackage{mathtools}
\usepackage{comment}
\usepackage{gensymb}
\usepackage{booktabs}
\usepackage{multirow}
\usepackage{array}
\newcolumntype{L}[1]{>{\raggedright\arraybackslash}p{#1}}
\newcolumntype{R}[1]{>{\raggedleft\arraybackslash}p{#1}}

\newcommand{\mbf}[1]{{\boldsymbol{\mathbf{#1}}}}
\renewcommand{\bm}{\mbf}

\usepackage{color}

\usepackage{url}
\usepackage{wrapfig}

\usepackage{titlesec}
\titlespacing*{\section}{1pt}{1pt}{1pt}
\titlespacing*{\subsection}{1pt}{1pt}{1pt}

\title{Scalable L\'evy Process Priors for Spectral Kernel Learning}

\author{
Phillip A.~Jang \quad Andrew E.~Loeb \quad Matthew B.~Davidow \quad Andrew Gordon Wilson \\
Cornell University
}

\begin{document} 

\maketitle

\begin{abstract}

Gaussian processes are rich distributions over functions, with generalization properties 
determined by a kernel function. When used for long-range extrapolation, predictions are 
particularly sensitive to the choice of kernel parameters.  It is therefore critical to account for 
kernel uncertainty in our predictive distributions.
We propose a distribution over kernels formed by modelling a spectral mixture density 
with a L\'evy process. 
The resulting distribution has support for all stationary covariances---including
the popular RBF, periodic, and Mat\'ern kernels---combined with inductive
biases which enable automatic and data efficient learning, long-range 
extrapolation, and state of the art predictive performance. The proposed model also
presents an approach to spectral regularization, as the L\'evy process introduces 
a sparsity-inducing prior over mixture components, allowing automatic selection 
over model order and pruning of extraneous components. We exploit the 
algebraic structure of the proposed process for $\mathcal{O}(n)$ training and 
$\mathcal{O}(1)$ predictions. We perform extrapolations having reasonable uncertainty 
estimates on several benchmarks, show that the proposed model can recover flexible 
ground truth covariances and that it is robust to errors in initialization.

\end{abstract}

\section{Introduction}
\label{sec: intro}
Gaussian processes (GPs) naturally give rise to a \emph{function space} view of modelling,
whereby we place a prior distribution over functions, and reason about the properties of 
likely functions under this prior \citep{rasmussen06}. Given data, we then infer a posterior 
distribution over functions to make predictions. The generalisation behavior of the Gaussian process
is determined by its prior support (which functions are a priori possible) 
and its inductive biases (which functions are a priori likely), which are in turn encoded by a 
kernel function. However, popular kernels, 
and even multiple kernel learning procedures, typically cannot extract highly expressive hidden 
representations, as was envisaged for neural networks \citep{mackay98, wilson2014thesis}.

To discover such representations, recent approaches have advocated building more 
expressive kernel functions.  For instance, spectral mixture kernels \citep{wilsonadams2013} 
were introduced for flexible kernel learning and extrapolation, by modelling a spectral
density with a scale-location mixture of Gaussians, with promising results.  However, 
\citet{wilsonadams2013} specify the number of mixture components by hand, and do not 
characterize uncertainty over the mixture hyperparameters.

As kernel functions become increasingly expressive and parametrized, it becomes natural 
to also adopt a function space view of kernel learning---to represent uncertainty
over the values of the kernel function, and to reflect the belief that the kernel does not 
have a simple form.  Just as we use Gaussian processes over 
functions to model data, we can apply the function space view a step further in a hierarchical 
model---with a prior distribution over kernels. 

In this paper, we introduce a scalable distribution over kernels by modelling a spectral 
density, the Fourier transform of a kernel, with a L\'evy process.  We consider both scale-location mixtures of Gaussians and Laplacians as basis functions for the L\'evy process, to induce a prior over kernels that gives rise to the sharply peaked spectral
densities that often occur in practice---providing a powerful inductive bias for kernel learning.
Moreover, this choice of basis functions allows our kernel function, conditioned on the L\'evy 
process, to be expressed in closed form.
This prior distribution over kernels also has 
support for \emph{all} stationary covariances---containing, for instance, any composition of the 
popular RBF, Mat\'{e}rn, rational quadratic, gamma-exponential, or spectral mixture 
kernels.   And unlike the spectral mixture representation in \citet{wilsonadams2013}, 
this proposed process prior allows for natural \emph{automatic} inference over the number of mixture 
components in the spectral density model.  Moreover, the priors implied 
by popular L\'evy processes such as the gamma process and symmetric $\alpha$-stable process result in 
even stronger complexity penalties than $\ell_1$ regularization, yielding sparse representations and removing 
mixture components which fit to noise. 

Conditioned on this distribution over kernels, we model data with a Gaussian process.
 To form a predictive distribution, we take a Bayesian model 
average of GP predictive distributions over a large set of possible kernel functions,
represented by the support of our prior over kernels, weighted by the posterior
probabilities of each of these kernels.  This procedure leads to a non-Gaussian 
heavy-tailed predictive distribution for modelling data. We develop a reversible jump 
MCMC (RJ-MCMC) scheme \citep{green1995} to infer the posterior distribution over kernels, 
including inference over the number of components in 
the L\'evy process expansion.  For scalability, we pursue a structured kernel 
interpolation \citep{wilsonnickisch2015} approach, in our case exploiting algebraic 
structure in the L\'evy process expansion, for $\mathcal{O}(n)$ inference and 
$\mathcal{O}(1)$ predictions, compared to the standard $\mathcal{O}(n^3)$ 
and $\mathcal{O}(n^2)$ computations for inference and predictions with 
Gaussian processes. Flexible distributions over kernels will
be especially valuable on large datasets, which often contain additional 
structure to learn rich statistical representations.

The key contributions of this paper are summarized as follows: 
\begin{enumerate}
\itemsep-0.20em
\item The first fully probabilistic approach to inference with spectral mixture kernels --- to incorporate kernel uncertainty into our predictive distributions, for a more realistic coverage of extrapolations. This feature is demonstrated in Section \ref{sec:spectralKernelLearning}.
\item Spectral regularization in spectral kernel learning. The L\'evy process prior acts as a sparsity-inducing prior on mixture components, automatically pruning extraneous components. This feature allows for automatic inference over model order, a key hyperparameter which must be hand tuned in the original spectral mixture kernel paper.
\item Reduced dependence on a good initialization, a key practical  improvement over the original spectral mixture kernel paper.
\item A conceptually natural and interpretable function space view of kernel learning.
\end{enumerate}

\section{Background}
\label{sec: background}
We provide a review of Gaussian and L\'evy processes as models for prior distributions over functions.

\subsection{Gaussian Processes}
\label{subsec: gps}
A stochastic process $f(\mathbf{x})$ is a Gaussian process (GP) if for any finite collection of inputs $X = \left\{\mathbf{x}_1, \cdots ,\mathbf{x}_n\right\} \subset \mathbb{R}^D$, the vector of function values $\left[ f(\mathbf{x}_1) , \cdots , f(\mathbf{x}_n) \right]^T$ is jointly Gaussian.

The distribution of a GP is completely determined by its mean function $m(\mathbf{x})$, and covariance kernel $k(\mathbf{x},\mathbf{x'})$.  A GP used to specify a distribution over functions is denoted as $f(\mathbf{x}) \sim \mathcal{GP}(m(\mathbf{x}),k(\mathbf{x},\mathbf{x'}))$, where $\mathbb{E}[f(x_i)] = m(\bm{x}_i)$ and $\text{cov}(f(\bm{x}),f(\bm{x}')) = k(\bm{x},\bm{x}')$. The generalization properties of the GP are encoded by the covariance kernel and its hyperparameters.

By exploiting properties of joint Gaussian variables, we can obtain closed form expressions for conditional mean and covariance functions of unobserved function values given observed function values. Given that $f(\mathbf{x})$ is observed at $n$ training inputs $X$ with values $\mathbf{f} = \left[ f(\mathbf{x}_1) , \cdots , f(\mathbf{x}_n) \right]^T$, the predictive distribution of the unobserved function values $\mathbf{f_*}$ at $n_*$ testing inputs $X_*$ is given by
\begin{align}
\mathbf{f_*}&|X_*,X,\theta \sim \mathcal{N}(\mathbf{\bar{f}_*}, \text{cov}(\mathbf{f}_*)), \\
\mathbf{\bar{f}_*} &= m_{X_*}+K_{X_*,X}K_{X,X}^{-1}(\mathbf{f}-m_X), \\
\text{cov}(\mathbf{f}_*) &= K_{X_*,X_*} - K_{X_*,X}K_{X,X}^{-1} K_{X,X_*}.
\end{align}

where $K_{X_*,X}$ for example denotes the $n_* \times n$ matrix of covariances evaluated at $X_*$ and $X$.

The popular radial basis function (RBF) kernel has the following form:
\begin{align}
k_{\text{RBF}}(\mathbf{x},\mathbf{x'}) = \exp(-0.5\left\|\mathbf{x}-\mathbf{x'}\right\|^2/\ell^2).
\end{align}

GPs with RBF kernels are limited in their expressiveness and act primarily as smoothing interpolators, because the only covariance structure they can learn from data is the length scale $\ell$, which determines how quickly covariance decays with distance. 

\citet{wilsonadams2013} introduce the more expressive \emph{spectral mixture} (SM) kernel capable of extracting more complex covariance structures than the RBF kernel, formed by placing a scale-location mixture of Gaussians in the spectrum of the covariance kernel. The RBF kernel in comparison can only model a single Gaussian centered at the origin in frequency (spectral) space.  

\subsection{L\'evy Processes}
\label{subsec:lpintro}

A stochastic process $\left\{L(\omega)\right\}_{\omega\in\mathbb{R}^+}$ is a \emph{L\'evy process} if it has stationary, independent increments and it is continuous in probability. In other words, $L$ must satisfy 

\begin{enumerate}
\item $L(0) = 0$,
\item $L(\omega_0), L(\omega_1) - L(\omega_0), \cdots , L(\omega_n) - L(\omega_{n-1})$ are independent $\forall \omega_0 \leq \omega_1 \leq \cdots \leq \omega_n$,
\item $L(\omega_2) - L(\omega_1) \stackrel{d}{=} L(\omega_2-\omega_1) \hspace{3mm} \forall \omega_2 \geq \omega_1$,
\item $\lim\limits_{h \to 0}\mathbb{P}(|L(\omega+h)-L(\omega)|\geq\varepsilon)=0 \hspace{3mm} \forall \varepsilon > 0 \ \forall \omega \geq 0$.
\end{enumerate}

\begin{wrapfigure}[13]{r}{0.5\textwidth}
\vspace{-.5cm}
\centering
\centerline{\includegraphics[width=0.5\textwidth]{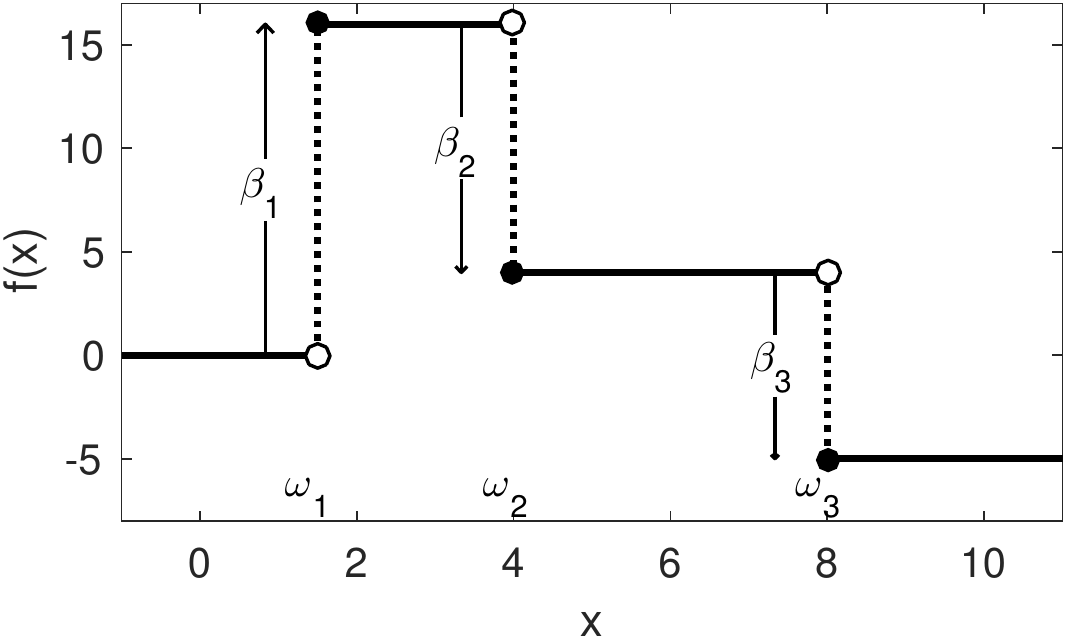}}
\caption{Annotated realization of a compound Poisson process, a special case of a L\'evy process. The $\omega_j$ represent jump locations, and $\beta_j$ represent jump magnitudes.}
\label{fig:LevyProcessModel}
\end{wrapfigure} 

By the L\'evy-Khintchine representation, the distribution of a (pure jump) L\'evy process is completely determined by its L\'evy measure. That is, the characteristic function of $L(\omega)$ is given by:
\begin{align*}
&\log\mathbb{E}[e^{iuL(\omega)}] =\\
& \omega\int_{\mathbb{R}^d\backslash\left\{0\right\}}\left(e^{iu \cdot \beta} - 1 - iu \cdot \beta 1_{|\beta| \leq 1}\right)\nu(d\beta).
\end{align*}
where the L\'evy measure $\nu(d\beta)$ is any $\sigma$-finite measure which satisfies the following integrability condition
\begin{equation*}
\int_{\mathbb{R}^d\backslash\left\{0\right\}}(1 \wedge \beta^2)\nu(d\beta) < \infty.
\end{equation*}

A L\'evy process can be viewed as a combination of a Brownian motion with drift and a superposition of independent Poisson processes with differing jump sizes $\beta$. The L\'evy measure $\nu(d\beta)$ determines the expected number of Poisson events per unit of time for any particular jump size $\beta$. The Brownian component of a L\'evy process will not be considered for this model.
For higher dimension input spaces $\omega \in \Omega$, one defines the more general notion of L\'evy random measure, which is also characterized by its L\'evy measure $\nu(d\beta d\omega)$ \citep{Wolpert} .
We will show that the sample realizations of L\'evy processes can be used to draw sample parameters for adaptive basis expansions.

\subsection{L\'evy Process Priors over Adaptive Expansions}
\label{subsec:lpadapt}
Suppose we wish to specify a prior over the class of adaptive expansions: $\left\{f : \mathcal{X} \rightarrow \mathbb{R} \;\middle|\; f(x)=\sum_{j=1}^J \beta_j \phi(x,\omega_j) \right\}$. Through a simple manipulation, we can rewrite $f(x)$ into the form of a stochastic integral:
\begin{equation*}
f(x)=\sum_{j=1}^J\beta_j\phi(x,\omega_j)=\sum_{j=1}^J\beta_j\int_{\Omega}\phi(x,\omega)\delta_{\omega_j}(\omega)d\omega=\int_\Omega \phi(x,\omega)\underbrace{\sum_{j=1}^J\beta_j\delta_{\omega_j}(\omega)d\omega}_{=dL(\omega)} .
\end{equation*}
Hence, by specifying a prior for the measure $L(\omega)$, we can simultaneously specify a prior for all of the parameters $\lbrace J,(\beta_1,\omega_1),...,(\beta_J,\omega_J)\rbrace$ of the expansion. L\'evy random measures provide a family of priors naturally suited for this purpose, as there is a one-to-one correspondence between the jump behavior of the L\'evy prior and the components of the expansion.

To illustrate this point, suppose the basis function parameters $\omega_j$ are one-dimensional and consider the  integral of $dL(\omega)$ from $0$ to $\omega$.
\begin{align*}
L(\omega) &= \int_0^{\omega}dL(\xi) = \int_0^{\omega}\sum_{j=1}^J\beta_j\delta_{\omega_j}(\xi)d\xi = \sum_{j=1}^J\beta_j 1_{\left[0,\omega\right]}(\omega_j).
\end{align*}
We see in Figure \ref{fig:LevyProcessModel} that $\sum_{j=1}^J\beta_j 1_{\left[0,\omega\right]}(\omega_j)$ resembles the sample path of a compound Poisson process, with the number of jumps $J$, jump sizes $\beta_j$, and jump locations $\omega_j$ corresponding to the number of basis functions, basis function weights, and basis function parameters respectively. We can use a compound Poisson process to define a prior over all such piecewise constant paths. More generally, we can use a L\'evy process to define a prior for $L(\omega)$.

Through the L\'evy-Khintchine representation, the jump behavior of the prior is characterized by a L\'evy measure $\nu(d\beta d\omega)$ which controls the mean number of Poisson events in every region of the parameter space, encoding the inductive biases of the model. As the number of parameters in this framework is random, we use a form of trans-dimensional reversible jump Markov chain Monte Carlo (RJ-MCMC) to sample the parameter space \citep{Green2003}.

Popular L\'evy processes such as the gamma process, symmetric gamma process, and the symmetric $\alpha$-stable process each possess desirable properties for different situations. The gamma process is able to produce strictly positive gamma distributed $\beta_j$ without transforming the output space. The symmetric gamma process can produce both positive and negative $\beta_j$, and according to \citet{Wolpert} can achieve nearly all the commonly used isotropic geostatistical covariance functions. The symmetric $\alpha$-stable process can produce heavy-tailed distributions for $\beta_j$ and is appropriate when one might expect the basis expansion to be dominated by a few heavily weighted functions. 

While one could dispense with L\'evy processes and place Gaussian or Laplace priors on $\beta_j$ to obtain $\ell_2$ or $\ell_1$ regularization on the expansions, respectively, a key benefit particular to these L\'evy process priors are that the implied priors on the coefficients yield even stronger complexity penalties than $\ell_1$ regularization. This property encourages sparsity in the expansions and permits scalability of our MCMC algorithm. Refer to the supplementary material for an illustration of the joint priors on coefficients, which exhibit concave contours in contrast to the convex elliptical and diamond contours of $\ell_2$ and $\ell_1$ regularization. Furthermore, in the log posterior for the L\'evy process there is a $\log(J!)$ complexity penalty term which further encourages sparsity in the expansions. Refer to \citet{clyde2007nonparametric} for further details.

\section{L\'evy Distributions over Kernels}
\label{sec: levydistkernels}
In this section, we motivate our choice of prior over kernel functions and describe how to generate samples from this prior distribution in practice.
\subsection{L\'evy Kernel Processes}
By Bochner's Theorem (\citeyear{bochner1959lectures}), a continuous stationary kernel can be represented as the Fourier dual of a spectral density: 
\begin{align}
k(\tau)=\int_{\mathbb{R}^D} S(s)e^{2\pi is^\top\tau}ds, \quad
S(s)=\int_{\mathbb{R}^D} k(\tau)e^{-2\pi is^\top\tau}d\tau.
\end{align}

Hence, the spectral density entirely characterizes a stationary kernel.  Therefore, it can be desirable to model the spectrum rather than the kernel, since we can then view kernel estimation through the lens of density estimation. 
In order to emulate the sharp peaks that characterize frequency spectra of natural phenomena, we model the spectral density with a location-scale mixture of Laplacian components:
\begin{align}
\phi_L(s,\omega_j) = \frac{\lambda_j}{2}e^{-\lambda_j |s-\chi_j|}, \quad
\omega_j \equiv (\chi_j,\lambda_j) \in  [0,f_{max}]\times\mathbb{R^+}. 
\label{eq:laplaceBasis}
\end{align}
Then the full specification of the symmetric spectral mixture is 
\begin{align}
\label{eq:spectralMixture}
S(s) = \frac{1}{2}\left[\tilde{S}(s) + \tilde{S}(-s)\right], \quad  
\tilde{S}(s) = \sum_{j=1}^J \beta_j \phi_L(s,\omega_j).
\end{align}
As Laplacian spikes have a closed form inverse Fourier transform, the spectral density $S(s)$ represents the following kernel function:
\begin{equation}
\label{eq:ift}
k(\tau)=\sum_{j=1}^J \beta_j \frac{\lambda_j^2}{\lambda_j^2+4\pi^2\tau^2}\text{cos}(2\pi\chi_j\tau).
\end{equation}

The parameters $J$, $\beta_j$, $\chi_j$, $\lambda_j$ can be interpreted through Eq.~\eqref{eq:ift}. 
The total number of terms to the mixture is $J$, while $\beta_j$ is the scale of the $j^{\text{th}}$ frequency contribution, $\chi_j$ is its central frequency, and $\lambda_j$ governs how rapidly the term decays (a high $\lambda$ results in confident, long-term periodic extrapolation). 

Other basis functions can be used in place of $\phi_L$ to model the spectrum as well. 
For example, if a Gaussian mixture is chosen, along with maximum likelihood estimation for the learning procedure, then we obtain the spectral mixture kernel \citep{wilsonadams2013}.

As the spectral density $S(s)$ takes the form of an adaptive expansion, we can define a L\'evy prior over all such densities and hence all corresponding kernels of the above form.
For a chosen basis function $\phi(s,\omega)$ and L\'evy measure $\nu(d\beta d\omega)$ we say that $k(\tau)$ is drawn from a \textbf{L\'evy kernel process (LKP)}, denoted as $k(\tau)~\sim~\mathcal{LKP}(\phi,\nu)$. \citet{Wolpert} discuss the necessary regularity conditions for $\phi$ and $\nu$. 
In summary, we propose the following hierarchical model over functions
\begin{align}
f(x)|k(\tau) \sim \mathcal{GP}(0,k(\tau)), \quad \tau = x-x', \quad k(\tau) \sim \mathcal{LKP}(\phi,\nu).
\end{align}

\begin{wrapfigure}[25]{r}{0.5\textwidth}
\vspace{-0.5cm}
\centering
\subfigure{\label{fig:a}\includegraphics[width=0.15\textwidth]{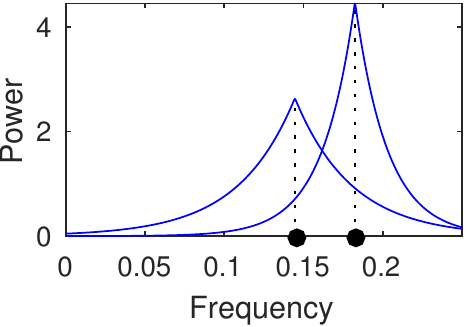}}
\subfigure{\label{fig:b}\includegraphics[width=0.15\textwidth]{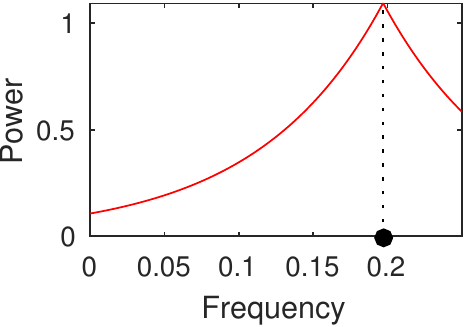}}
\subfigure{\label{fig:c}\includegraphics[width=0.15\textwidth]{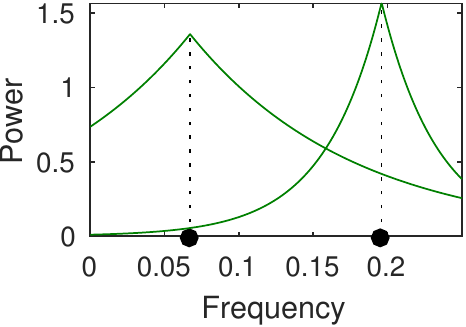}}
\subfigure{\label{fig:d}\includegraphics[width=0.5\textwidth]{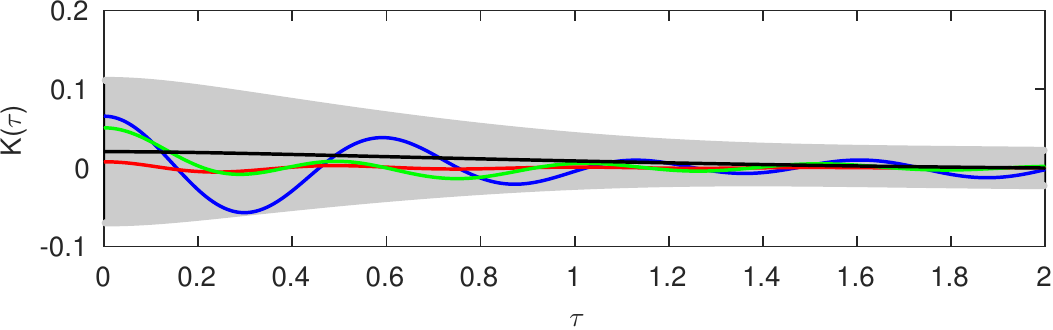}}
\subfigure{\label{fig:e}\includegraphics[width=0.5\textwidth]{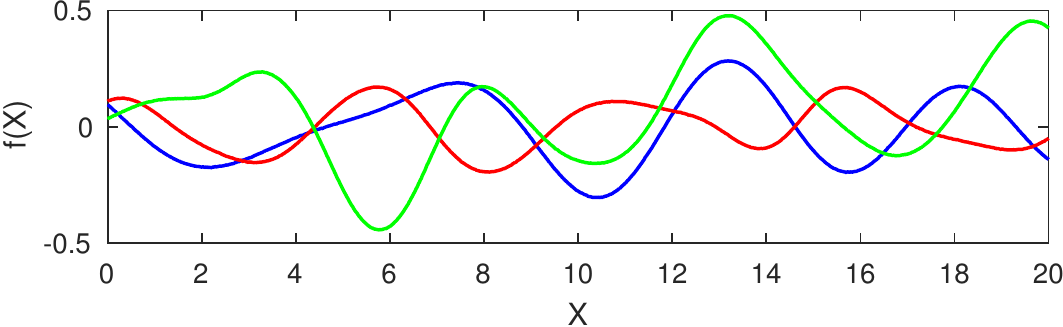}}
\caption{Samples from a L\'evy kernel mixture prior distribution. (top) Three spectra with Laplace components drawn from a L\'evy process prior. (middle) The corresponding stationary covariance kernel functions and the prior mean with two standard deviations of the model, as determined by 10,000 samples. (bottom) GP samples with the respective covariance kernel functions.}
\label{fig:priorSamples}
\end{wrapfigure}

Figure \ref{fig:priorSamples} shows three samples from the L\'evy process specified through Eq.~\eqref{eq:spectralMixture} and their corresponding covariance kernels. 
We also show one GP realization for each of the kernel functions.
By placing a L\'evy process prior over spectral densities, we induce a L\'evy kernel process prior over stationary covariance functions.

\subsection{Sampling L\'evy Priors}
We now discuss how to generate samples from the L\'evy kernel process in practice. 
In short, the kernel parameters are drawn according to $\{J,\{(\beta_j,\omega_j)\}_{j=1}^J\} \sim \text{L\'evy}(\nu(d\beta d\omega))$, and then Eq.~\eqref{eq:ift} is used to evaluate $k \sim \mathcal{LKP}(\phi_L,\nu)$ at values of $\tau$. 

Recall from Section \ref{subsec:lpadapt} that the choice of L\'evy measure $\nu$ is completely determined by the choice of the corresponding L\'evy process and vice versa. 
Though the processes mentioned there produce sample paths with infinitely many jumps (and cannot be sampled directly), almost all jumps are infinitesimally small, and therefore these processes can be approximated in $L^2$ by a compound Poisson process with a jump size distribution truncated by $\varepsilon$.

Once the desired L\'evy process is chosen and the truncation bound is set, the basis expansion parameters are generated by drawing $J \sim \text{Poisson}(\nu_\varepsilon^+),$ and then drawing $J$ i.i.d. samples $\beta_1, \cdots, \beta_J \sim \pi_\beta(d\beta),$ and $J$ i.i.d. samples $\omega_1, \cdots, \omega_J \sim \pi_\omega(d\omega)$. Refer to the supplementary material for $L^2$ error bounds and formulas for $\nu_\varepsilon^+ = \nu_\varepsilon(\mathbb{R} \times \Omega)$ for the gamma, symmetric gamma, and symmetric $\alpha$-stable processes. 

The form of $\pi_\beta(\beta_j)$ also depends on the choice of L\'evy process and can be found in the supplementary material, with further details in \citet{Wolpert}.
We choose to draw $\chi$ from an uninformed uniform prior over a reasonable range in the frequency domain, and $\lambda$ from a gamma distribution, $\lambda \sim \text{Gamma}(a_\lambda,b_\lambda)$. 
The choices for $a_\lambda$, $b_\lambda$, and the frequency limits are left as hyperparameters, which can have their own hyperprior distributions.
After drawing the $3J$ values that specify a L\'evy process realization, the corresponding covariance function can be evaluated through the analytical expression for the inverse Fourier transform (e.g. Eq.~\eqref{eq:ift} for Laplacian frequency mixture components).

\section{Scalable Inference}
\label{sec: inference}
Given observed data $\mathcal{D} = \lbrace x_i,y_i\rbrace_{i=1}^N $, we wish to infer $p(y(x_*)|D, x_*)$ over some test set of inputs $x_*$ for interpolation and extrapolation. We model observations $y(x)$ with a hierarchical model:
\begin{align}
\label{Eq: Model Hierarchy}
y(x)|f(x) &= f(x) + \varepsilon(x), \hspace{5mm} \varepsilon(x) \stackrel{\text{iid}}{\sim} N(0,\sigma^2), \\
f(x)|k(\tau) &\sim \mathcal{GP}(0,k(\tau)), \hspace{5mm} \tau = x-x', \\
k(\tau) &\sim \mathcal{LKP}(\phi,\nu).
\end{align}
Computing the posterior distributions by marginalizing over the LKP will yield a heavy-tailed non-Gaussian process for $y(x_*)=y_*$ given by an infinite Gaussian mixture model:
\begin{align}
p(y_*|\mathcal{D}) = \int p(y_*|k,\mathcal{D})p(k|\mathcal{D})dk 
\label{Eq:MCMC Sum}
                      \approx \frac{1}{H}\sum_{h=1}^H p(y_* | k_h), \hspace{3mm} k_h \sim p(k | \mathcal{D}).
\end{align}

We compute this approximating sum using $H$ RJ-MCMC samples \citep{Green2003}. Each sample draws a kernel from the posterior $k_h \sim p(k | \mathcal{D})$ distribution. Each sample of $k_h$ enables us to draw a sample from the posterior predictive distribution $p(y_* | \mathcal{D})$, from which we can estimate the predictive mean and variance.

Although we have chosen a Gaussian observation model in Eq.~\eqref{Eq: Model Hierarchy} (conditioned on $f(x)$), all of the inference procedures we have introduced here would also apply to non-Gaussian likelihoods, such as for Poisson processes with Gaussian process intensity functions, or classification.

The sum in Eq.~\eqref{Eq:MCMC Sum} requires drawing kernels from the distribution $p(k|\mathcal{D})$. This is a difficult distribution to approximate, particularly because there is not a fixed number of parameters as $J$ varies.  We employ RJ-MCMC, which extends the capability of conventional MCMC to allow sequential samples of different dimensions to be drawn \citep{Green2003}. 
Thus, a posterior distribution is not limited to coefficients and other parameters of a fixed basis expansion, but can represent a changing number of basis functions, as required by the description of L\'evy processes described in the previous section. 
Indeed, RJ-MCMC can be used to automatically learn the appropriate number of basis functions in an expansion.
In the case of spectral kernel learning, inferring the number of basis functions corresponds to automatically learning the important frequency contributions to a GP kernel, which can lead to new interpretable insights into our data.

\subsection{Initialization Considerations}
\label{subsec:initialization}

The choice of an initialization procedure is often an important practical consideration for machine learning tasks due to severe multimodality in a likelihood surface \citep{neal1996}.
In many cases, however, we find that spectral kernel learning with RJ-MCMC can automatically learn salient frequency contributions with a simple initialization, such as a uniform covering over a broad range of frequencies with many sharp peaks. The frequencies which are not important in describing the data are quickly attenuated or removed within RJ-MCMC learning. Typically only a few hundred RJ-MCMC iterations are needed to discover the salient frequencies in this way.  

\citet{wilson2014thesis} proposes an alternative structured approach to initialization in previous spectral kernel modelling work.  First, pass the (squared) data through a Fourier transform to obtain an empirical spectral density, which can be treated as observed.  Next, fit the empirical spectral density using a standard Gaussian mixture density estimation procedure, assuming a fixed number of mixture components.  Then, use the learned parameters of the Gaussian mixture as an initialization of the spectral mixture kernel hyperparameters, for Gaussian process marginal likelihood optimization.
We observe successful adaptation of this procedure to our L\'evy process method, replacing the approximation with Laplacian mixture terms and using the result to initialize RJ-MCMC.

\subsection{Scalability}
\label{sec:scalability}
As with other GP based kernel methods, the computational bottleneck lies in the evaluation of the log marginal likelihood during MCMC, which requires computing $(K_{X,X}+\sigma^2I)^{-1}y$ and $\log|K_{X,X} + \sigma^2I|$ for an $n \times n$ kernel matrix $K_{X,X}$ evaluated at the $n$ training points $X$. 
A direct approach through computing the Cholesky decomposition of the kernel matrix requires $\mathcal{O}(n^3)$ computations and $\mathcal{O}(n^2)$ storage, restricting the size of training sets to $\mathcal{O}(10^4)$.
Furthermore, this computation must be performed at every iteration of RJ-MCMC, compounding standard computational constraints.

However, this bottleneck can be readily overcome through the Structured Kernel Interpolation approach introduced in \citet{wilsonnickisch2015}, which approximates the kernel matrix as $\tilde{K}_{X,X'}=M_X K_{Z,Z} M_{X'}^\top$ for an exact kernel matrix $K_{Z,Z}$ evaluated on a much smaller set of $m \ll n$ inducing points, and a sparse interpolation matrix $M_X$ which facilitates fast computations. 
The calculation reduces to $\mathcal{O}(n+g(m))$ computations and $\mathcal{O}(n+g(m))$ storage.  As described in \citet{wilsonnickisch2015}, we can impose Toeplitz structure on $K_{Z,Z}$ for $g(m) = m \log m$, allowing our RJ-MCMC procedure to train on massive datasets.

\section{Experiments}
\label{sec: experiments}

We conduct four experiments in total. In order to motivate our model for kernel learning in later experiments, we first demonstrate the ability of a L\'evy process to recover---through direct regression---an observed noise-contaminated spectrum that is characteristic of sharply peaked naturally occurring spectra. In the second experiment we demonstrate the robustness of our RJ-MCMC sampler by automatically recovering the generative frequencies of a known kernel, even in presence of significant noise contamination and poor initializations.  In the third experiment we demonstrate the ability of our method to infer the spectrum of airline passenger data, to perform long-range extrapolations on real data, and to demonstrate the utility of accounting for uncertainty in the kernel.  In the final experiment we demonstrate the scalability of our method through training the model on a 100,000 data point sound waveform. Code is available at \url{https://github.com/pjang23/levy-spectral-kernel-learning}.

\subsection{Explicit Spectrum Modelling}
\label{subsec: directSpectrum}

\begin{wrapfigure}[13]{r}{0.5\textwidth}
\vspace{-1.2cm}
\centering
\centerline{\includegraphics[width=0.5\textwidth]{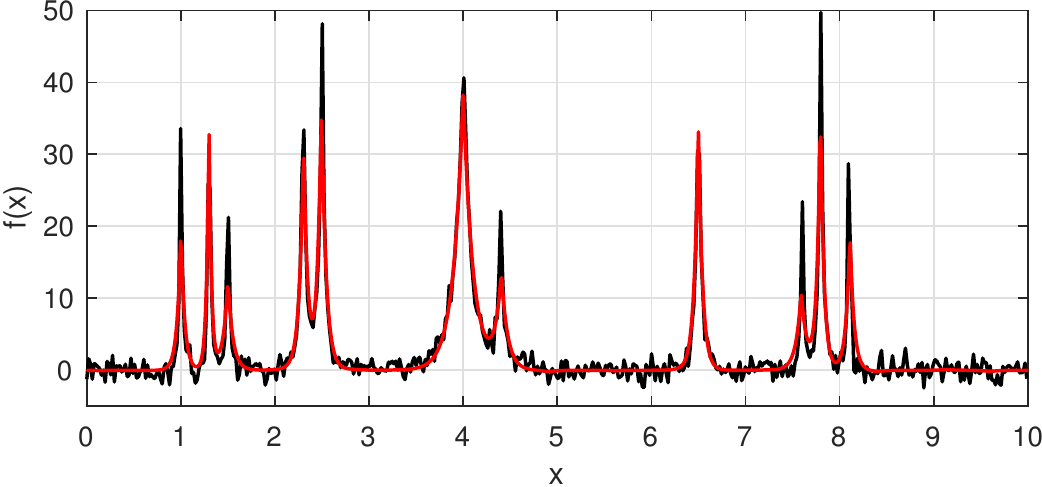}}
\vspace{-0.2cm}
\caption{L\'evy process regression on a noisy test function (black). The fit (red) captures the locations and scales of each spike while ignoring noise, but falls slightly short at its modes since the black spikes are parameterized as $(1+|x|)^{-4}$ rather than Laplacian.}
\label{fig:explicitSpectrumFit}
\end{wrapfigure} 

We begin by applying a L\'evy process directly for function modelling (known as LARK regression), with inference as described in \citet{Wolpert}, and Laplacian basis functions.
We choose an out of class test function proposed by \citet{Donoho} that is standard in wavelet literature. 
The spatially inhomogeneous function is defined to represent spectral densities that arise in scientific and engineering applications. Gaussian i.i.d. noise is added to give a signal-to-noise ratio of 7, to be consistent with previous studies of the test function \citet{Wolpert}. 

The noisy test function and LARK regression fit are shown in Figure \ref{fig:explicitSpectrumFit}. 
The synthetic spectrum is well characterized by the L\'evy process, with no ``false positive'' basis function terms fitting the noise owing to the strong regularization properties of the L\'evy prior. 
By contrast, GP regression with an RBF kernel learns a length scale of 0.07 through maximum marginal likelihood training: 
the Gaussian process posterior can fit the sharp peaks in the test function only if it also overfits to the additive noise.

The point of this experiment is to show that the L\'evy process with Laplacian basis functions forms a natural prior over spectral densities.  In other words, samples from this prior will typically look like the types of spectra that occur in practice.  Thus, this process will have a powerful inductive bias when used for kernel learning, which we explore in the next experiments.

\newpage

\subsection{Ground Truth Recovery}
\label{subsec: groundTruth}
\vspace{-0.2cm}

\begin{wrapfigure}[22]{r}{0.5\textwidth}
\vspace{-1.2cm}
\centering
\subfigure{\label{fig:f}\includegraphics[width=0.22\textwidth]{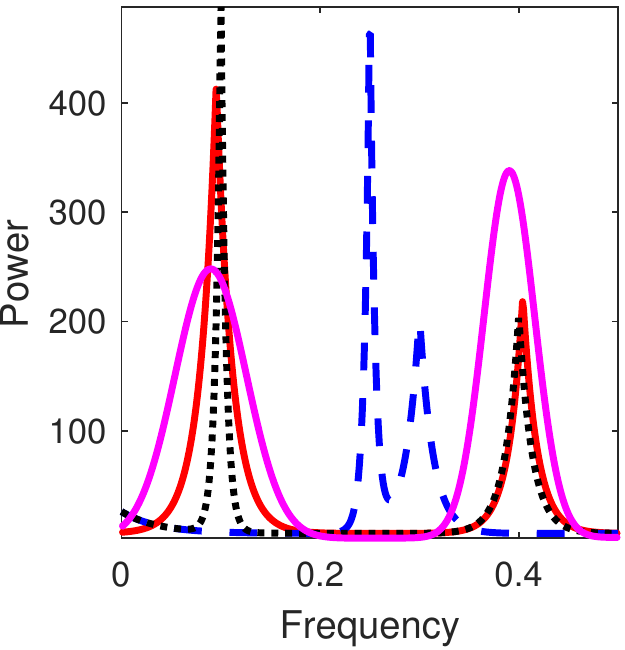}}
\subfigure{\label{fig:g}\includegraphics[width=0.23\textwidth]{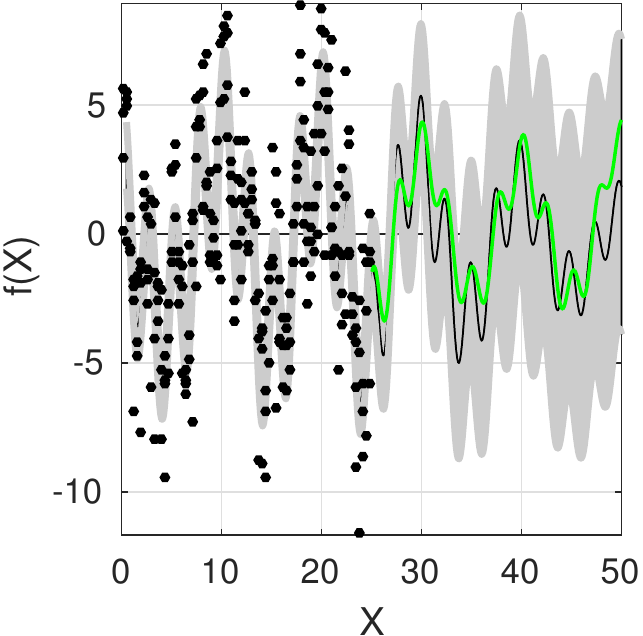}}
\caption{Ground truth recovery of known frequency components. (left) The spectrum of the Gaussian process that was used to generate the noisy training data is shown in black. From these noisy data and the erroneous spectral initialization shown in dashed blue, the maximum a posteriori estimate of the spectral density (over 1000 RJ-MCMC steps) is shown in red.
A SM kernel also identifies the salient frequencies, but with broader support, shown in magenta. (right) Noisy training data are shown with a scatterplot, with withheld testing data shown in green.  The learned posterior predictive distribution (mean in black, with 95\% credible set in grey) captures the test data.}
\label{fig:groundTruthRecovery}
\end{wrapfigure}

We next demonstrate the ability of our method to recover the generative frequencies of a known kernel and its robustness to noise and poor initializations. Data are generated from a GP with a kernel having two spectral Laplacian peaks, and partitioned into training and testing sets containing 256 points each. Moreover, the training data are contaminated with i.i.d. Gaussian noise (signal-to-noise ratio of 85\%). 

Based on these observed training data (depicted as black dots in Figure \ref{fig:groundTruthRecovery}, right), we estimate the kernel of the Gaussian process by inferring its spectral density (Figure \ref{fig:groundTruthRecovery}, left) using 
1000 RJ-MCMC iterations.
The empirical spectrum initialization described in section \ref{subsec:initialization} results in the discovery of the two generative frequencies. Critically, we can also recover these salient frequencies \emph{even with a very poor initialization}, as shown in Figure \ref{fig:groundTruthRecovery} (left).  

For comparison, we also train a Gaussian SM kernel, initializing based on the empirical spectrum. The resulting kernel spectrum (Figure \ref{fig:groundTruthRecovery}, magenta curve) does recover the salient frequencies, though with less confidence and higher overhead than even a poor initialization and spectral kernel learning with RJ-MCMC.

\subsection{Spectral Kernel Learning for Long-Range Extrapolation}
\label{sec:spectralKernelLearning}
\vspace{-0.2cm}

\begin{wrapfigure}[18]{r}{0.5\textwidth}
\vspace{-0.5cm}
\centering
\centerline{\includegraphics[width=0.5\textwidth]{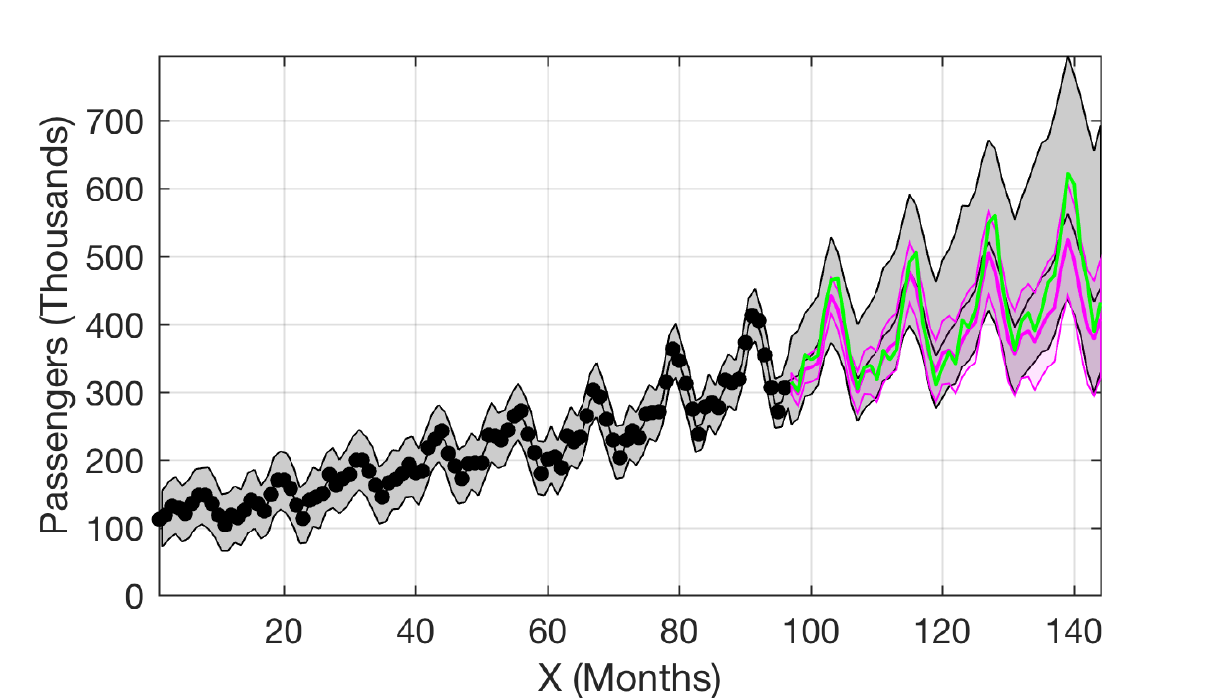}}
\vspace{-0.2cm}
\caption{Learning of Airline passenger data. Training data is scatter plotted, with withheld testing data shown in green.  The learned posterior distribution with the proposed approach (mean in black, with 95\% credible set in grey) captures the periodicity and the rising trend in the test data. The analogous 95\% interval using a GP with a SM kernel is illustrated in magenta.}
\label{fig:airlineFit}
\end{wrapfigure} 

We next demonstrate the ability of our method to perform long-range extrapolation on real data. Figure \ref{fig:airlineFit} shows a time series of monthly airline passenger data from 1949 to 1961 \citep{hyndman2005}. The data show a long-term rising trend as well as a short term seasonal waveform, and an absence of white noise artifacts.

As with \citet{wilsonadams2013}, the first 96 monthly data points are used to train the model and the last 48 months (4 years) are withheld as testing data, indicated in green. With an initialization from the empirical spectrum and 2500 RJ-MCMC steps, the model is able to automatically learn the necessary frequencies and the shape of the spectral density to capture both the rising trend and the seasonal waveform, allowing for accurate long-range extrapolations without pre-specifying the number of model components in advance.

This experiment also demonstrates the impact of accounting for uncertainty in the kernel, as the withheld data often appears near or crosses the upper bound of the 95\% predictive bands of the SM fit, whereas our model yields wider and more conservative predictive bands that wholly capture the test data. As the SM extrapolations are highly sensitive to the choice of parameter values, fixing the parameters of the kernel will yield overconfident predictions. The L\'evy process prior allows us to account for a range of possible kernel parameters so we can achieve a more realistically broad coverage of possible extrapolations.

\begin{figure}
\centering
\includegraphics[width=\textwidth]{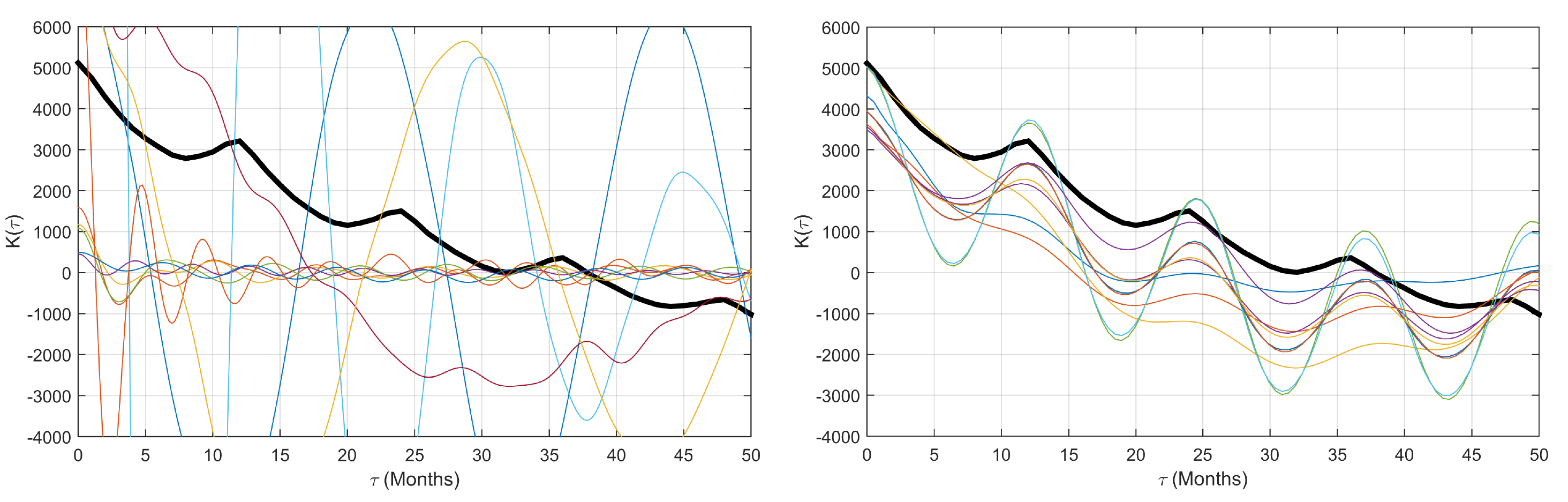}
\vspace{-0.7cm}
\caption{Covariance function draws from the kernel prior (left) and posterior (right) distributions, with the empirical covariance function shown in black. After RJ-MCMC, the covariance distribution centers upon the correct frequencies and order of magnitude.}
\vspace{-0.7cm}
\label{fig:covSamples}
\end{figure}

Note that the L\'evy process over spectral densities induces a prior over kernel functions.  Figure \ref{fig:covSamples}
 shows a side-by-side comparison of covariance function draws from the prior and posterior distributions over kernels.  We see that sample covariance functions from the prior vary quite significantly, but are concentrated in the posterior, with movement towards the empirical covariance function.

\subsection{Scalability Demonstration}
\label{subsec: soundmodel}
\vspace{-0.2cm}

\begin{wrapfigure}[18]{r}{0.45\textwidth}
\vspace{-0.7cm}
\centering
\centerline{\includegraphics[width=0.5\textwidth]{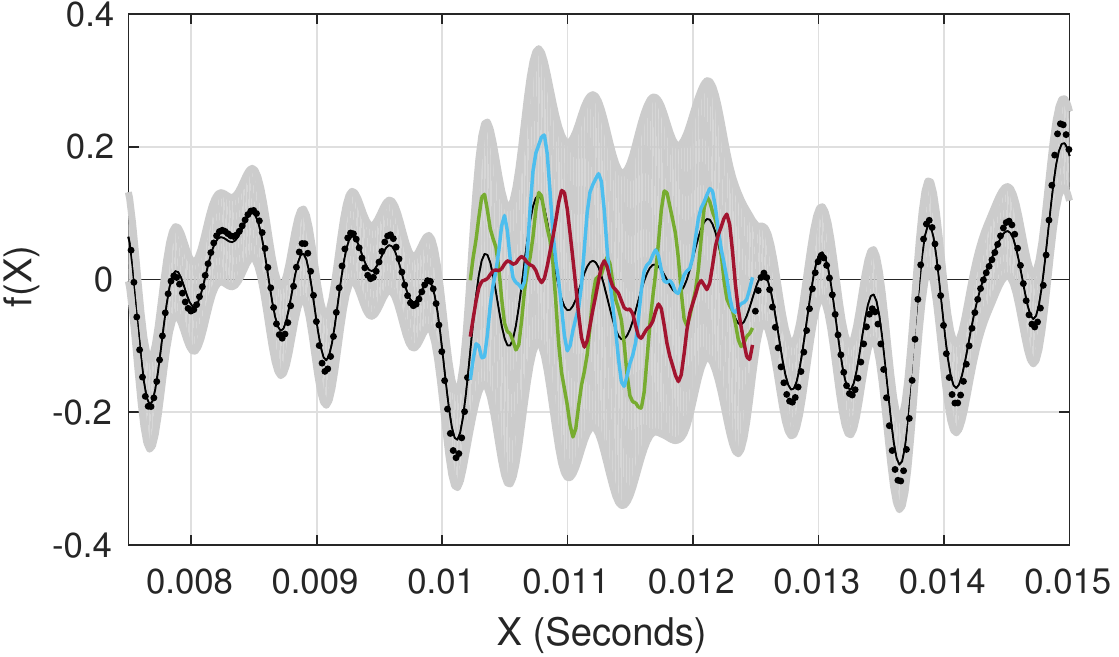}}
\caption{Learning of a natural sound texture. A close-up of the training interval is displayed with the true waveform data scatter plotted. The learned posterior distribution (mean in black, with 95\% credible set in grey) retains the periodicity of the signal within the corrupted interval. Three samples are drawn from the posterior distribution. }
\label{fig:soundInfilling}
\end{wrapfigure}

A flexible and fully Bayesian approach to kernel learning can come with some additional computational overhead.  Here we demonstrate the scalability that is achieved through the integration of SKI \citep{wilsonnickisch2015} with our L\'evy process model.

We consider a 100,000 data point waveform, taken from the field of natural sound modelling \citep{Turner}.
A L\'evy kernel process is trained on a sound texture sample of howling wind with the middle 10\% removed. Training involved initialization from the signal empirical covariance and 500 RJ-MCMC samples, and took less than one hour using an Intel i7 3.4 GHz CPU and 8 GB of memory. Four distinct mixture components in the model were automatically identified through the RJ-MCMC procedure.
The learned kernel is then used for GP infilling with 900 training points, taken by down-sampling the training data, which is then applied to the original 44,100 Hz natural sound file for infilling.

The GP posterior distribution over the region of interest is shown in Figure \ref{fig:soundInfilling}, along with sample realizations, which appear to capture the qualitative behavior of the waveform. This experiment demonstrates the applicability of our proposed kernel learning method to large datasets, and shows promise for extensions to higher dimensional data.

\section{Discussion}
\label{sec: discussion}

We introduced a distribution over covariance kernel functions that is well suited for modelling quasi-periodic data. 
We have shown how to place a L\'evy process prior over the spectral density of a stationary kernel.  The resulting hierarchical model allows the incorporation of kernel uncertainty into the predictive distribution. Through the spectral regularization properties of L\'evy process priors, we found that our trans-dimensional sampling procedure is suitable for automatically performing inference over model order, and is robust over initialization strategies.  Finally, we incorporated structured kernel interpolation into our training and inference procedures for linear time scalability, enabling experiments on large datasets.  The key advances over conventional spectral mixture kernels are in 
being able to interpretably and automatically discover the number of mixture components, and in representing uncertainty
over the kernel.  Here, we considered one dimensional inputs and stationary processes to most clearly elucidate the key properties of L\'evy kernel processes.  However, one could generalize this process to multidimensional non-stationary kernel learning by jointly inferring properties of transformations over inputs alongside the kernel hyperparameters.  Alternatively, one could consider neural networks as basis functions in the L\'evy process, inferring distributions over the parameters of the network and the numbers of basis functions as a step towards automating neural network architecture construction.  

\textbf{Acknowledgements.} This work is supported in part by the Natural Sciences and Engineering Research Council of Canada (PGS-D 502888) and the National Science Foundation DGE 1144153 and IIS-1563887 awards.

\bibliography{mbibnew,references} 
\bibliographystyle{icml2017}

\newpage

\section{Supplementary Materials}

\subsection{Sampling Levy Process Priors}
\label{section:Priors}
The following formulas in this section are taken from \citet{Wolpert} for reference.

Suppose the hyperparameters $\theta$ of the prior distributions for $J,\beta,\omega$, are drawn from a hyperprior distribution, $\pi_\theta(d\theta)$. Then in order to sample the L\'evy prior, the follow steps are taken: 
\begin{align*}
\theta &\sim \pi_\theta(d\theta) \\
J|\theta &\sim \text{Po}(\nu_\varepsilon^+), \hspace{5mm} \nu_\varepsilon^+ \equiv \nu_\varepsilon( \mathbb{R} \times \Omega) \\
\lbrace(\beta_j,\omega_j)\rbrace_{j=1}^J | J,\theta &\overset{\text{i.i.d.}}{\sim} \pi_\beta(\beta_j)d\beta_j\pi_\omega(d\omega_j)
\end{align*}

The formulas for $\nu_\varepsilon^+$ and $\pi_\beta$ are determined by the specific choice of L\'evy process and are given below. For computational purposes, the $\beta_j$'s are truncated at $|\beta_j\eta|>\varepsilon$ for a Poisson approximation to the true L\'evy process, and $\mathbb{E}|\mathcal{L}[\phi]-\mathcal{L}_\varepsilon[\phi]|^2$ represents the $L^2$ error of the approximation for a given basis function $\phi$. Below, $E_1(z) = \int_z^{\infty}t^{-1}e^{-t}dt$.

\subsubsection{Gamma Process}

$$ J \sim \text{Po}(\nu_\varepsilon^+), \hspace{1cm} \nu_\varepsilon^+=\gamma|\Omega|E_1(\varepsilon)$$

$$\beta_j \overset{\text{i.i.d.}}{\sim} \pi_\beta(\beta_j)d\beta_j, \hspace{1cm} \pi_\beta(\beta_j)=\frac{\beta_j^{-1}e^{-\beta_j\eta}}{E_1(\varepsilon)}\bf{1}_{\lbrace \beta_j\eta>\varepsilon \rbrace}$$

$$\mathbb{E}|\mathcal{L}[\phi]-\mathcal{L}_\varepsilon[\phi]|^2 = \gamma\eta^{-2}\|\phi\|_2^2[1-(1+\varepsilon)e^{-\varepsilon}]$$

\subsubsection{Symmetric Gamma Process}

$$ J \sim \text{Po}(\nu_\varepsilon^+), \hspace{1cm} \nu_\varepsilon^+=2\gamma|\Omega|E_1(\varepsilon)$$

$$\beta_j \overset{\text{i.i.d.}}{\sim} \pi_\beta(\beta_j)d\beta_j, \hspace{1cm} \pi_\beta(\beta_j)=\frac{|\beta_j|^{-1}e^{-|\beta_j|\eta}}{2E_1(\varepsilon)}\bf{1}_{\lbrace |\beta_j\eta|>\varepsilon \rbrace}$$

$$\mathbb{E}|\mathcal{L}[\phi]-\mathcal{L}_\varepsilon[\phi]|^2 = 2\gamma\eta^{-2}\|\phi\|_2^2[1-(1+\varepsilon)e^{-\varepsilon}]$$

\subsubsection{Symmetric $\alpha$-Stable Process}
$$ J \sim \text{Po}(\nu_\varepsilon^+), \hspace{1cm} \nu_\varepsilon^+=\gamma|\Omega|\frac{2}{\pi}\Gamma(\alpha)\text{sin}\left(\frac{\pi\alpha}{2}\right)\varepsilon^{-\alpha}$$

$$\beta_j \overset{\text{i.i.d.}}{\sim} \pi_\beta(\beta_j)d\beta_j, \hspace{1cm} \pi_\beta(\beta_j)=\frac{\alpha \varepsilon^\alpha}{2 \eta^\alpha} |\beta_j|^{-\alpha-1} \bf{1}_{\lbrace |\beta_j\eta|>\varepsilon \rbrace}$$

$$\mathbb{E}|\mathcal{L}[\phi]-\mathcal{L}_\varepsilon[\phi]|^2 = 2\gamma\eta^{-2}\|\phi\|_2^2\left[\frac{\Gamma(\alpha+1)}{\pi(2-\alpha)}\sin\left(\frac{\pi\alpha}{2}\right)\varepsilon^{2-\alpha}\right]$$

\newpage

\subsection{Sparsity Inducing Properties of L\'evy Priors}
\begin{figure}[h!]
\centering
\includegraphics[width=0.9\textwidth]{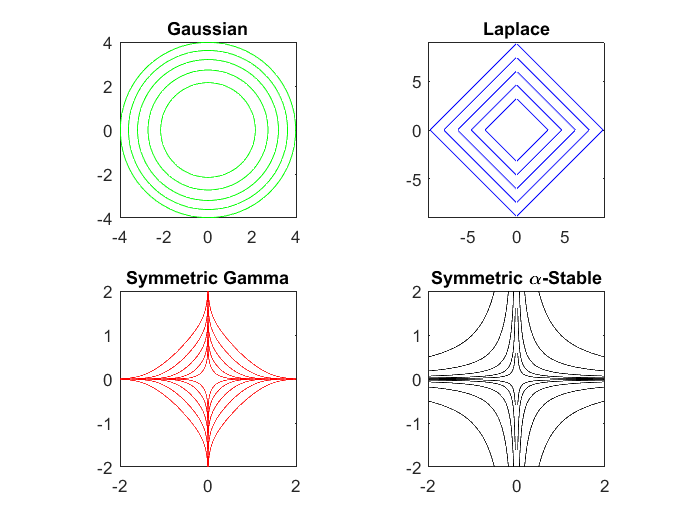}
\caption{\label{fig:contour} Contour plot of the joint probability density function of two $\beta$ draws under different priors.}
\end{figure}

Figure \ref{fig:contour} illustrates the contours of the joint distribution for two independent draws of $\beta$ under different priors $\pi_\beta(d\beta)$. The contours for the gamma process would be taken from the upper-right quadrant of those for the symmetric gamma process.

Gaussian and Laplace priors on $\beta$ result in $\ell_2$ and $\ell_1$ regularization respectively. The L\'evy processes in contrast yield inward curving contours, leading to a sparsity inducing effect similar to $\ell_p$ regularization with $p<1$. Intuitively, this discourages simultaneous large values of $\beta$ more strongly than $\ell_1$ regularization unless the added basis functions significantly improve the fit.

\subsection{Initialization and Hyperparameter Tuning}
\label{sec: initialization}

\begin{figure}[h!]
\centering
\includegraphics[width=0.85\textwidth]{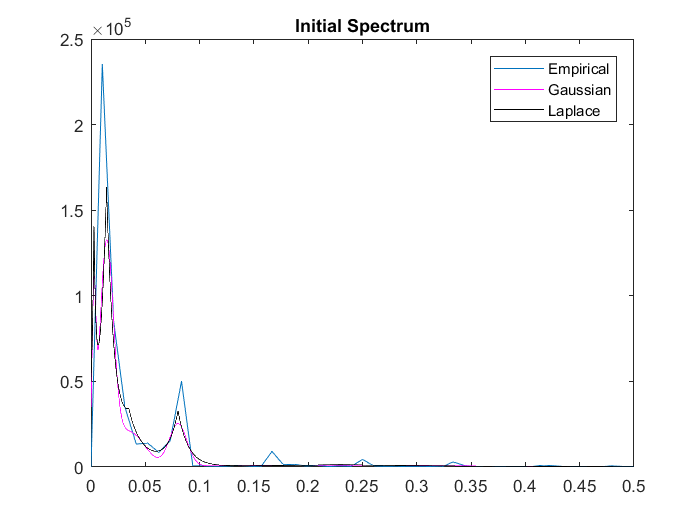}
\caption{\label{fig:FitDemean}Initial Spectrum Fit}
\end{figure}

Initialization and hyperparameter tuning can be automated by fitting the empirical spectrum of the data. It is done in the following steps:

\begin{enumerate}
\item If needed, de-mean the training data by subtracting a deterministic mean function such as the sample mean or best fit line. Doing so will eliminate large peaks at the origin which dominate the rest of the spectrum. The de-meaned training data $\{y_j\}_{j=1}^n$ will be the input for RJ-MCMC.
\item Compute the empirical spectral density $S_{emp}(s) = \frac{2}{n}\left|\sum_{j=1}^n y_j e^{-2\pi is(j-1)}\right|^2, s \in [0,0.5]$. In MATLAB, this is calculated as the first $\lfloor \frac{n}{2} \rfloor$ entries from 2*abs(fft(y)).\^{}2/n;
\item Sample the empirical spectral density and fit a Gaussian mixture with $J_0$ components to the sampled data. A good initial guess for $J_0$ can be done by examining the number of peaks in the empirical spectrum.
$$
S_{\text{Gaussian}}(s) = \sum_{j=1}^{J_0} \alpha_j \frac{1}{\sqrt{2\pi\sigma_j^2}}e^{-\frac{(s-\chi_j)^2}{2\sigma_j^2}}
$$
\item Keep the frequencies $\chi_j$ from the Gaussian fit, and using least squares, fit a Laplacian basis function to each individual Gaussian component. For each $j$, one could minimize the following objective over a sample grid of points ${s_k}$ in $[-3\sigma_j,3\sigma_j]$ 
$$
\min_{\lambda_j,\beta_j} \sum_{s_k} \left[\beta_j \frac{\lambda_j}{2} e^{-\lambda_j|s_k|} - \alpha_j \frac{1}{\sqrt{2\pi\sigma_j^2}} e^{-\frac{s_k^2}{2\sigma_j^2}}\right]^2
$$
\item Form the initial spectrum with the fitted parameters
$$
S_{\text{initial}}(s) = \sum_{j=1}^{J_0} \beta_j \frac{\lambda_j}{2}e^{-\lambda_j |s-\chi_j|} 
$$The initial spectrum fit for the airline data is shown in Figure \ref{fig:FitDemean}.

\item Tune the hyperparameters:
	\begin{itemize}
	\item $\lambda$ is modelled with prior $\text{Gamma}(a_\lambda,b_\lambda)$, so $a_\lambda$ and $b_\lambda$ can be estimated by maximum likelihood on the $\lambda$ parameters of the initial spectrum.
	\item $\eta^{-1} \sim \text{Gamma}(a_\eta,b_\eta)$ controls the expected value of coefficients $\beta_j$. For basis functions which integrate to 1, the sum of $\beta_j$'s is equal to the total area underneath the spectrum, which by Parseval's identity represents total variance of the data. Hence the sample variance of the training data can be used as an upper bound on coefficient values, and $a_\eta$ and $b_\eta$ can be set accordingly.
	\item $\gamma \sim \text{Gamma}(a_\gamma,b_\gamma)$ is proportional to the expected number of basis functions as shown in Section \ref{section:Priors} and controls the sparsity of the expansions. $a_\gamma$ and $b_\gamma$ can be set to cover a range of values which encourage sparsity.
	\item For the symmetric $\alpha$-stable process, $0 < \alpha < 2$ controls the heaviness of the tails in the distribution for $\beta_j$ with smaller values of $\alpha$ yielding heavier tails. $\alpha$ can be set by maximum likelihood on the initial $\beta_j$'s.
	\item $\varepsilon$ can be set based on $L^2$ truncation errors as described in Section \ref{section:Priors}.
	\end{itemize}
\end{enumerate}

\end{document}